\documentclass{CVM}

\usepackage{amsmath,amssymb,amsfonts}
\usepackage{amsthm, mathrsfs}
\usepackage{multirow}
\usepackage{color}
\usepackage{float}
\usepackage{makecell}
\usepackage{hyperref}
\usepackage{url}
\usepackage{algorithmic}
\usepackage{graphicx}
\usepackage{subfigure}
\usepackage{textcomp}
\usepackage{bbding}
\usepackage{CJKutf8}

\CVMsetup{
type      = {Review Article},
doi       = {s41095-0xx-xxxx-x},
title     = {A Survey on Deep Learning-based Spatio-temporal Action Detection},
author    = {Peng Wang$^{1}$, Fanwei Zeng$^{2}$, and Yuntao Qian$^{1}$\cor{}\\
},
runauthor = {F. A. Author, S. B. Author, T. C. Author},
abstract  = {
  Spatio-temporal action detection (STAD) aims to classify the actions present in a video and localize them in space and time. It has become a particularly active area of research in computer vision because of its explosively emerging real-world applications, such as autonomous driving, visual surveillance, entertainment, \textit{etc}. Many efforts have been devoted in recent years to building a robust and effective framework for STAD. This paper provides a comprehensive review of the state-of-the-art deep learning-based methods for STAD. Firstly, a taxonomy is developed to organize these methods. Next, the linking algorithms, which aim to associate the frame- or clip-level detection results together to form action tubes, are reviewed. Then, the commonly used benchmark datasets and evaluation metrics are introduced, and the performance of state-of-the-art models is compared. At last, this paper is concluded, and a set of potential research directions of STAD are discussed.
},
keywords  = {Computer vision; deep learning; spatio-temporal action detection},
copyright = {The Author(s)},
}

\begin{document}
\begin{CJK}{UTF8}{gbsn}

\maketitle

\begin{figure}[b] \vskip -4mm
\small\renewcommand\arraystretch{1.3}
    \begin{tabular}{p{80.5mm}} \bottomrule\\ \end{tabular}
    \vskip -4.5mm \noindent \setlength{\tabcolsep}{1pt}
    \begin{tabular}{p{3.5mm}p{80mm}}
$1\quad $ & College of Computer Science, Zhejiang University, Hangzhou, Zhejiang 310007, China. E-mail: P. Wang, pengwang18@zju.edu.cn; Y.-T. Qian, ytqian@zju.edu.cn\cor{}.\\
$2\quad $ & Ant Group, Hangzhou, Zhejiang 310007, China. E-mail: fanwei.zfw@antgroup.com.\\
&\hspace{-5mm} Manuscript received: 2022-01-01; accepted: 2022-01-01\vspace{-2mm}
\end{tabular} \vspace {-7mm}
\end{figure}

\section{Introduction}

\subsection{Motivation}

\begin{figure*}[t]
  \centering
  \includegraphics[width=6.9in]{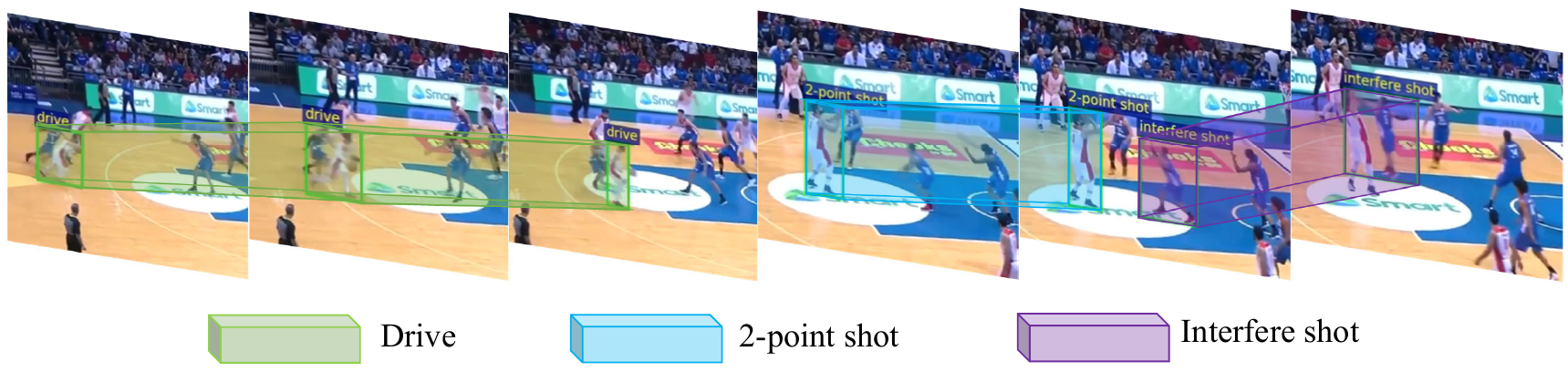}
  \caption{Spatio-temporal action detection classifies and localizes actions in space and time. Images are from MultiSports~\cite{li2021multisports} dataset.}
  \label{fig_task}
\end{figure*}

Spatio-temporal action detection (STAD), which aims to localize actions in space and time in long untrimmed videos as well as predict action categories (see Figure~\ref{fig_task}), is an essential and challenging task in video understanding. Because of its fundamental role in many applications, such as surveillance, sports analysis, robotics and self-driving cars, it has attracted a lot of attention and been actively researched in computer vision communities.

Before the prevalence of deep learning, traditional STAD usually involves a sliding window approach~\cite{Tian2013SpatiotemporalDP, Yuan2011DiscriminativeVP, siva2011weakly, lan2011discriminative, ke2007event}, such as deformable part models~\cite{Tian2013SpatiotemporalDP}, branch and bound approach~\cite{Yuan2011DiscriminativeVP}, \textit{etc}. Recently, with the significant development of deep neural networks, especially the convolutional neural network~\cite{LeCun1998GradientbasedLA} and transformer \cite{vaswani2017attention}, many significant advances in STAD have been achieved. These deep learning-based methods outperform the traditional algorithms by large margins and continue to improve the state-of-the-art.

With the rapid progress in deep learning-based STAD, a multitude of literature is being produced in this field. However, the survey of this area is scarce. Vahdani \textit{et al.}~\cite{Vahdani2021DeepLA} reviewed the action detection task in untrimmed videos with the emphasis on temporal action detection that aims to detect the start and end of action instances, and only a brief introduction was given to STAD. Bhoi \textit{et al.}~\cite{Bhoi2019SpatiotemporalAR} conducted a survey of STAD, but they focused on introducing traditional methods, and only three early deep learning-based approaches were reviewed. Thus, the recent development of this area is missing in their survey. Overall, there is a lack of comprehensive and in-depth survey in deep learning-based STAD, though it is highly needed for further research in this field. 

In response, we provide the first review that systematically introduces the most recent advances in deep learning-based STAD for interested researchers who would like to enter this ever-changing field and experts who want to compare STAD models and datasets. We collect abundant resources in this field, including state-of-the-art models, linking algorithms, benchmark datasets, performance comparison, \textit{etc}. We hope this survey will facilitate the advancement of STAD. 


\subsection{Organization}

The rest of this survey is organized as follows. Section~\ref{sec_taxonomy} clarifies the categorization of deep learning-based STAD methods. Section~\ref{sec_link} reviews how the detection results in each frame or clip are linked together. Section~\ref{sec_dataset} summarizes the benchmark datasets and evaluation metrics and compares the performance among state-of-the-art methods. Section~\ref{sec_future} points out a set of future directions. Finally, Section~\ref{sec_conclusion} concludes this survey.

\section{Taxonomy and Methods}  \label{sec_taxonomy}

In this section, we first describe the STAD problem formulation. Then we illustrate the taxonomy of deep learning-based STAD methods. Finally, we review these methods in detail in Subsection~\ref{subsec_frame} and \ref{subsec_clip}.

\subsection{Problem Formulation}   \label{subsec_formulation}

Given a video of $T$ frames $\{I_t\}_{t=1..T}$, the spatio-temporal action detection task, sometimes termed action localization or event detection, determines what actions occur in this video, and when and where they occur. That is to say, STAD models should output the action label $c_i \in \mathcal{C}$ ($\mathcal{C}$ is the set of action classes), as well as a set of bounding boxes (or regions) $\{R^i_t\}_{t=t_b..t_e}$, where $t_b$ is the beginning and $t_e$ is the end of the predicted action $c_i$, and $R^i_t$ is the detected region in frame $I_t$. Notably, action recognition and temporal action detection are closely related to STAD, but they only determine what or when action occurs (see Table~\ref{table_three_tasks}). Thus, they are not as challenging as STAD.  

\begin{table}[t]
\center
\caption{Comparison of action recognition, temporal action detection (TAD), and spatio-temporal action detection (STAD)}
  \begin{tabular}{p{2.5cm}<{\centering} p{1.0cm}<{\centering} p{1.7cm}<{\centering} p{1.5cm}<{\centering}}
  \bottomrule
                            & Action class    & The start and end of action  & Actor localization \\
  \hline
  Action recognition        & \Checkmark      & \XSolidBrush                 & \XSolidBrush       \\
  TAD                       & \Checkmark      & \Checkmark                   & \XSolidBrush       \\
  STAD                      & \Checkmark      & \Checkmark                   & \Checkmark         \\
  \toprule
  \multicolumn{4}{l}{"\Checkmark" means "need to be determined".} \\
  \multicolumn{4}{l}{"\XSolidBrush" means "need not to be determined".} \\
\end{tabular}
\label{table_three_tasks}
\end{table}

\begin{figure}[h!t]
  \centering
  \includegraphics[width=3.3in]{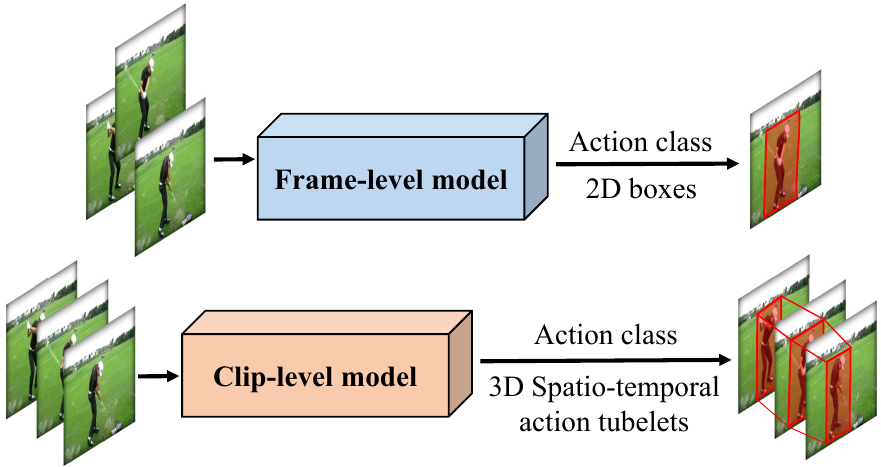}
  \caption{An illustration of the output of the frame-level model and clip-level model}
  \label{fig_frame_clip}
\end{figure}

\subsection{Taxonomy of Deep Learning-based STAD Methods}

\begin{figure*}[t]
  \centering
  \includegraphics[width=6.8in]{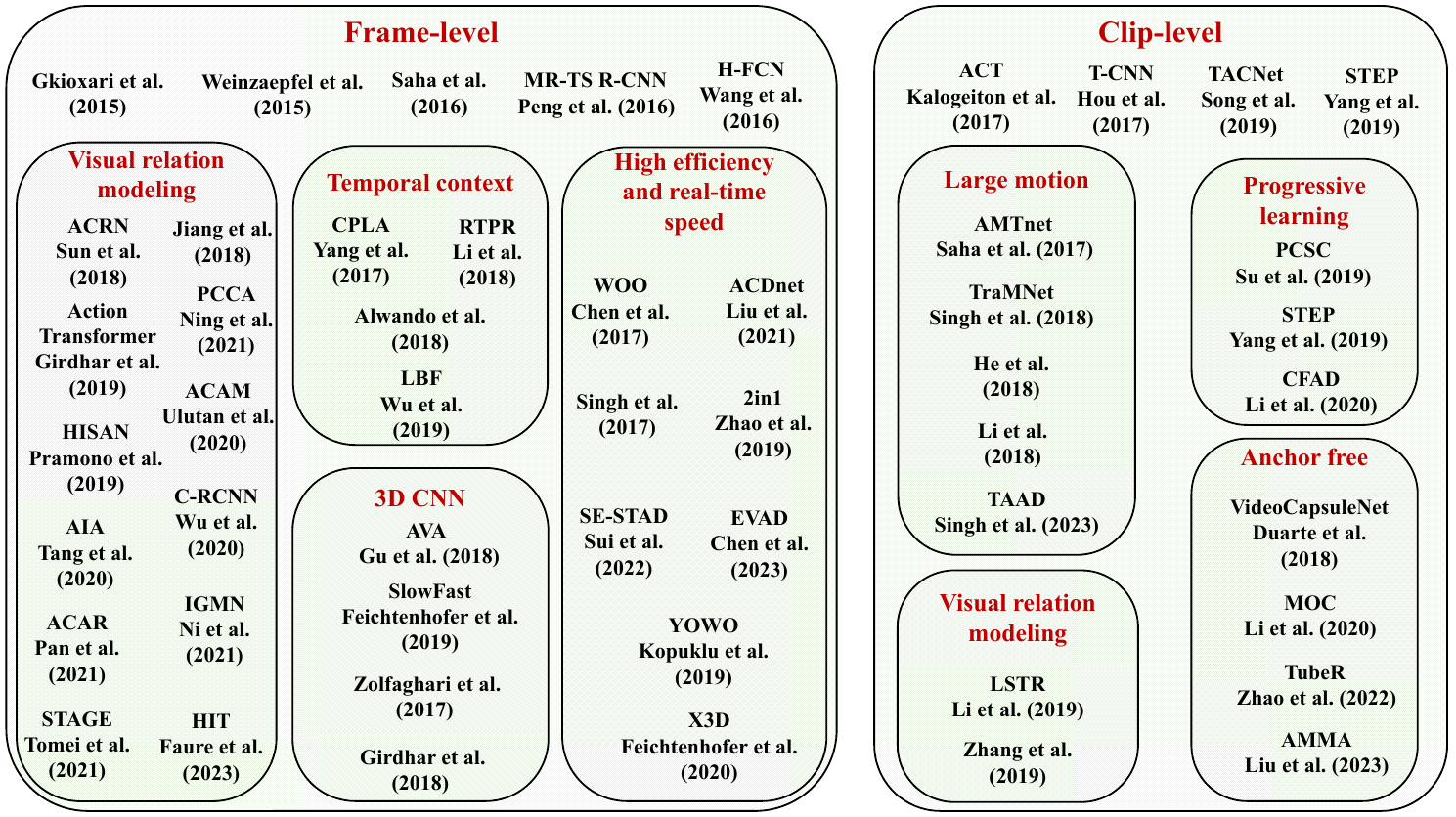}
  \caption{Taxonomy of deep learning-based STAD models}
  \label{fig_taxonomy_ring}
\end{figure*}

A wide variety of deep learning-based STAD methods have been proposed so far. These models can be mainly divided into two categories: frame-level and clip-level. Whereas frame-level models predict 2D bounding boxes for a frame, the clip-level models predict 3D spatio-temporal tubelets for a clip. Fig.~\ref{fig_frame_clip} illustrates the output paradigms of these two categories of models. To provide an in-depth review of deep learning-based STAD models, we subdivide the frame-level and clip-level models according to the motivation of model design. Since the motivation behind a model often reflects researchers' insights in the STAD field, we hope this categorization can provide a clear way to sort out the methods in this field and give researchers inspiration. Fig.~\ref{fig_taxonomy_ring} illustrates our taxonomy and some representative models.

\begin{figure}[t]
  \centering
  \includegraphics[width=3.3in]{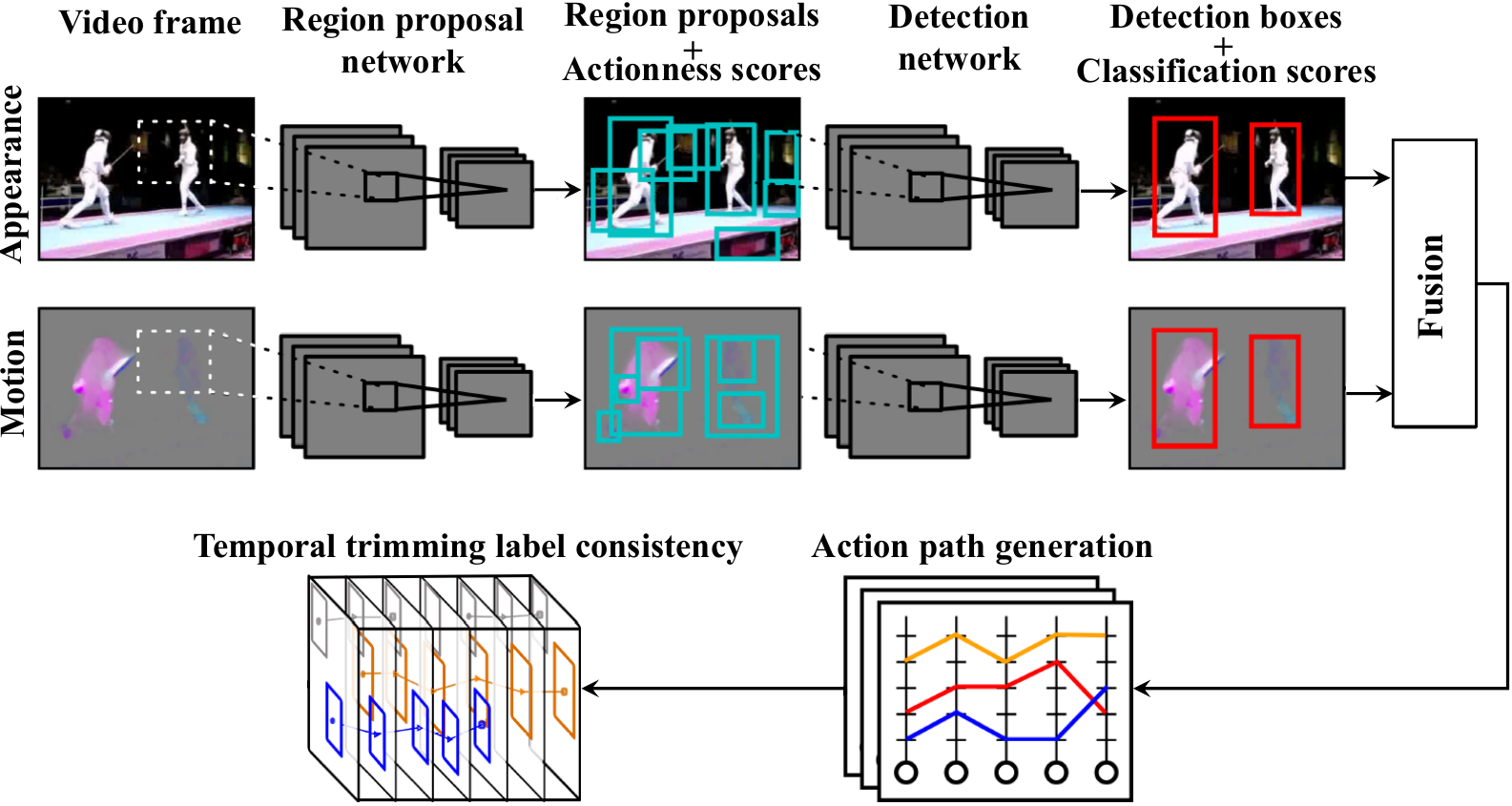}
  \caption{Flowchart of the method proposed by Saha \textit{et al.} (2016)~\cite{Saha2016DeepLF}.}
  \label{fig_2016_saha}
\end{figure}

\subsection{Frame-level Methods}   \label{subsec_frame}

The past decade has witnessed the dramatic development of object detection, with numerous models being proposed, such as R-CNN~\cite{girshick2014rich}, Faster R-CNN~\cite{ren2015faster}, YOLO~\cite{redmon2016you}, SSD~\cite{liu2016ssd}, \textit{etc}. Inspired by these advancements, many researchers seek to generalize the object detection models to the STAD field. A straightforward generalization is to regard the STAD in the video as a set of 2D image detections. Concretely, one applies an action detector at each frame independently to produce frame-level 2D bounding boxes. Then, 3D action proposals  (\textit{i.e.}, action tubes) are generated by associating these frame-level detection results using linking or tracking algorithms. 

Along this research line, Gkioxari \textit{et al.}~\cite{Gkioxari2014FindingAT} proposed a model based on R-CNN~\cite{girshick2014rich}. To incorporate both appearance and motion cues, they adopted two-stream architecture, with a spatial stream operating on the RGB frames and a temporal stream on the optical flow. Weinzaepfel \textit{et al.}~\cite{Weinzaepfel2015LearningTT} presented a method that extracts a set of candidate regions at the frame level via EdgeBoxes~\cite{zitnick2014edge} and then tracks high-scoring proposals throughout the video using a tracking-by-detection approach. They determined the temporal extension of actions by using a sliding-window approach. Driven by the huge success of Faster R-CNN~\cite{ren2015faster}, Saha \textit{et al.}~\cite{Saha2016DeepLF} introduced the first model that replaces the unsupervised region proposal algorithms with the region proposal network (RPN) for STAD, as shown in Fig~\ref{fig_2016_saha}. They passed RGB and optical-flow images to two separate region proposal networks to output detection boxes and action class scores. These appearance and motion-based detections were fused and linked up to generate class-specific action tubes. The effectiveness of RPN on STAD was further verified by Peng \textit{et al.}~\cite{Peng2016MultiregionTR}, where they found that compared to other proposal generalization methods, RPN achieves consistently better results with higher inter-section-over-union (IoU) score. Moreover, they experimentally found that stacking optical flow over several frames can improve frame-level action detection. Besides these proposal-based action detection methods, Wang \textit{et al.}~\cite{Wang2016ActionnessEU} introduced H-FCN, a method that detects actions based on the estimated actionness maps, where actionness means the likelihood of containing a generic action instance at a specific location of an image.

\textbf{Temporal context.} The above STAD methods treat each frame independently, ignoring the temporal contextual relationships. To overcome this problem, Yang \textit{et al.}~\cite{Yang2017SpatioTemporalAD} proposed a cascade proposal and location anticipation model, named CPLA, which is capable of inferring the movement trend of action occurrences between two frames. It uses detected bounding boxes on frame $I_t$ to infer the corresponding boxes on frame $I_{t+k}$, where $k$ is the anticipation gap. As a follow-up work, Li \textit{et al.}~\cite{Li2018RecurrentTP} presented an approach that models the temporal correlations of proposals between two consecutive frames to predict movements. Whereas \cite{Yang2017SpatioTemporalAD} requires running RPN at each frame, \cite{Li2018RecurrentTP} only runs RPN at the first frame of a video. In another work, Alwando \textit{et al.}~\cite{alwando2018cnn} proposed a video localization refinement scheme to iteratively rectify the potentially inaccurate bounding boxes by exploiting the
temporal consistency between adjacent frames. Wu \textit{et al.}~\cite{wu2019long} presented an insight that actions can become clear when relating the target frame to the long-range context. Therefore, they proposed a long-term feature bank (LFB) that provides long-term
supportive information to video models, enabling them to better understand the present.

\begin{figure}[t]
  \centering
  \includegraphics[width=3.3in]{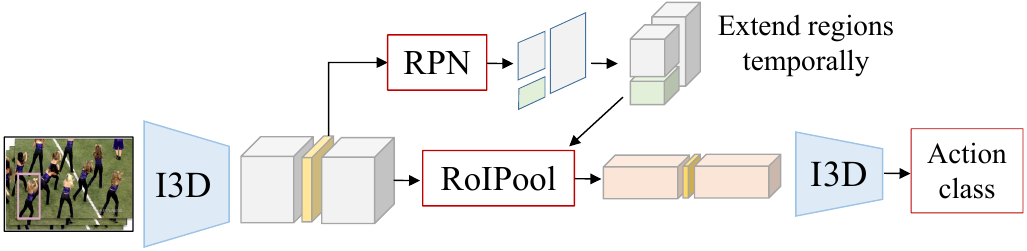}
  \caption{Architecture of the model proposed by Girdhar \textit{et al.} (2018)~\cite{girdhar2018better}.}
  \label{fig_2018_girdhar}
\end{figure}

\textbf{3D CNN.} Apart from the methods mentioned above that capture the motion characteristics in videos via optical flow, another group of work adopts 3D convolutional neural networks (3D CNNs)~\cite{ji20123d} to extract the motion information encoded in multiple adjacent frames. Gu \textit{et al.}~\cite{Gu2018AVAAV} proposed to combine inflated 3D convolutional network (I3D)~\cite{carreira2017quo} with Faster R-CNN~\cite{ren2015faster}. They first fed input frames to the I3D model to extract 3D feature maps and used ResNet-50 model on the keyframe to generate region proposals. Then, they extended ROI Pooling to 3D by applying the 2D ROI Pooling at the same spatial location over all time steps so that the spatio-temporal feature for each proposal was obtained, which was then fed into the classifier for action labeling. Girdhar \textit{et al.}~\cite{girdhar2018better} improved the performance of \cite{Gu2018AVAAV} by applying the I3D features for both proposal generation and classification, see Fig.~\ref{fig_2018_girdhar}. They passed a video clip through the first few blocks of I3D to get a video representation. Then, the center frame representation was used to predict potential `person' regions using an RPN. The proposals were extended in time by replicating and used to extract a feature map for the region using ROI pooling. The feature map was then classified into different actions using the two last I3D blocks. Zolfaghari \textit{et al.}~\cite{zolfaghari2017chained} proposed a network architecture that integrates three visual cues for action recognition and detection: pose, motion, and raw images. They introduced a Markov chain model that adds cues successively to integrate the features. Feichtenhofer \textit{et al.}~\cite{feichtenhofer2019slowfast} presented a SlowFast network involving two pathways: a slow pathway and a fast pathway. Whereas the former operates at a low frame rate to capture spatial semantics, the latter operates at a high frame rate to extract motion information at fine temporal resolution. The fast pathway is intentionally made very lightweight by reducing its channel capacity. With these innovative designs, SlowFast achieved strong performance for STAD in videos.

\textbf{High efficiency and real-time speed}. Current stage-of-the-art STAD models usually push performance by larger backbone network and more complex architecture and training procedures, thereby suffering from heavy computational burden and low efficiency. To move beyond such limitations, some researchers strived to design efficient and real-time models. Feichtenhofer \textit{et al.}~\cite{feichtenhofer2020x3d} presented an X3D network that was constructed by progressively expanding a tiny 2D architecture along multiple network axes, including space, time, width, and depth. X3D achieved state-of-the-art performance while requiring 5.5$\times$ fewer parameters for similar accuracy as previous work. Liu \textit{et al.}~\cite{liu2021acdnet} introduced ACDnet that performs feature approximation at most frames by exploiting the temporal coherence between successive video frames, instead of conducting the time-consuming CNN feature extraction.

\begin{figure}[t]
  \centering
  \includegraphics[width=3.2in]{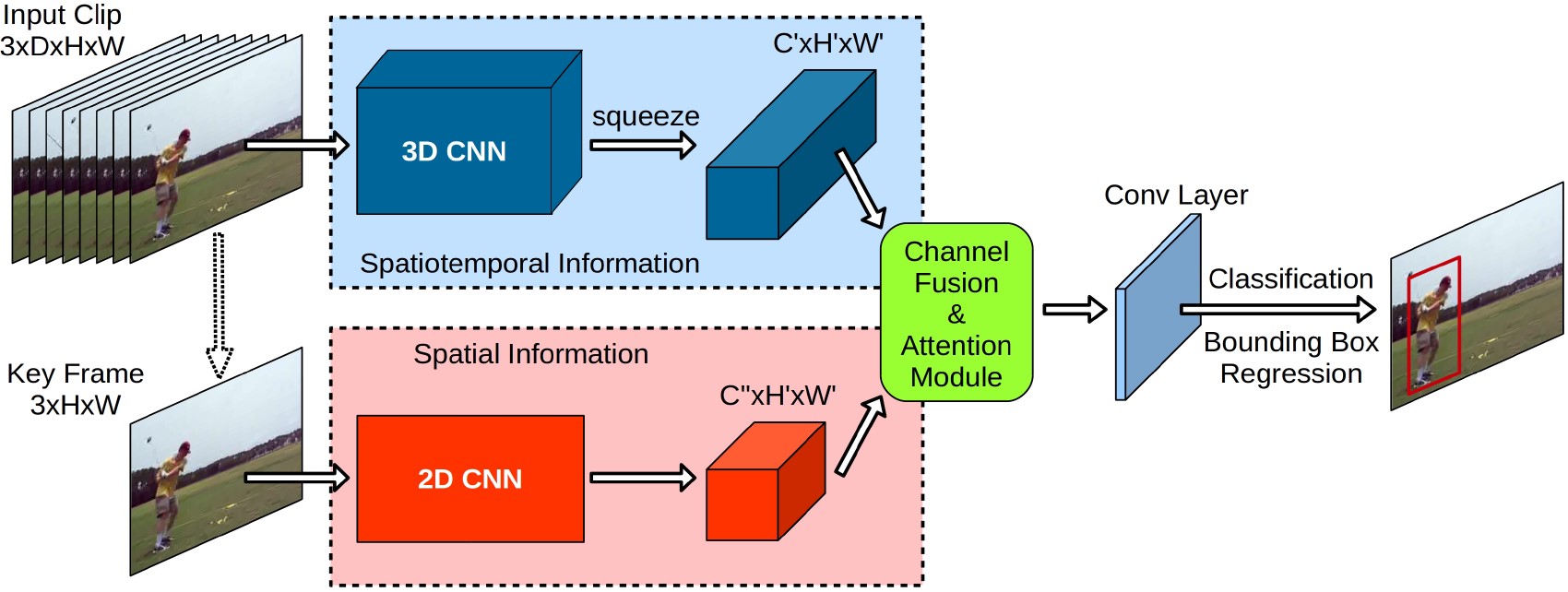}
  \caption{Network architecture of YOWO~\cite{Kpkl2019YouOW}.}
  \label{fig_YOWO}
\end{figure}

Zhao \textit{et al.}~\cite{zhao2019dance} argued that the two-stream detection network based on RGB and flow provides state-of-the-art accuracy at the expense of a large model size and heavy computation. Hence, they proposed embedding RGB and optical flow into a single stream network. They introduced a motion modulation layer to leverage optical flow to modulate RGB features. Their network, called 2in1, has half the computation and parameters of a two-stream equivalent while obtaining better action detection accuracy. 

Singh \textit{et al.}~\cite{Singh2016OnlineRM} proposed a method that can perform STAD in an online setting and at real-time speed, owing to two major developments: 1) they adopted real-time SSD~\cite{liu2016ssd} CNNs to detect actors; 2) they designed an online algorithm to incrementally construct action tubes from the frame-level detections (see Section~\ref{sec_link_frame_level}).

Inspired by the outstanding real-time performance of YOLO~\cite{redmon2016you}, Kopuklu \textit{et al.} put forward a STAD counterpart network, \textit{i.e.}, YOWO~\cite{Kpkl2019YouOW}, which replaces the detect-then-classify two-stage paradigm with the end-to-end one-stage paradigm, as shown in Fig.~\ref{fig_YOWO}. It outputs the regions of interest (ROIs) and the corresponding class prediction simultaneously. Albeit efficient, YOWO adopts two backbones, \textit{i.e.}, a 3D CNN for extracting spatio-temporal information and a 2D model for extracting spatial features. To combat this limitation, Chen \textit{et al.}~\cite{Chen2021WatchOO} presented a single unified network (\textit{i.e.}, WOO) that only adopts one backbone for both actor localization and action classification. Compared to two-backbone STAD models, WOO reduces the model complexity by over 50\%. However, it suffers from a performance drop. To improve the performance, they designed an extra attention-based embedding interaction module to obtain more discriminative features. As a follow-up work, SE-STAD~\cite{Sui2022ASA} achieves stronger performances than WOO~\cite{Chen2021WatchOO} with better training strategies and FCOS~\cite{tian2019fcos} as object detector. Most recently, Chen \textit{et al.} proposed a framework for efficient video action detection (EVAD)~\cite{chen2023efficient} based on vanilla visual transformers (ViTs). They reduced computational costs by dropping out the non-keframe tokens and enhanced the model performance by refining scene context.


\begin{figure}[t]
  \centering
  \includegraphics[width=3.2in]{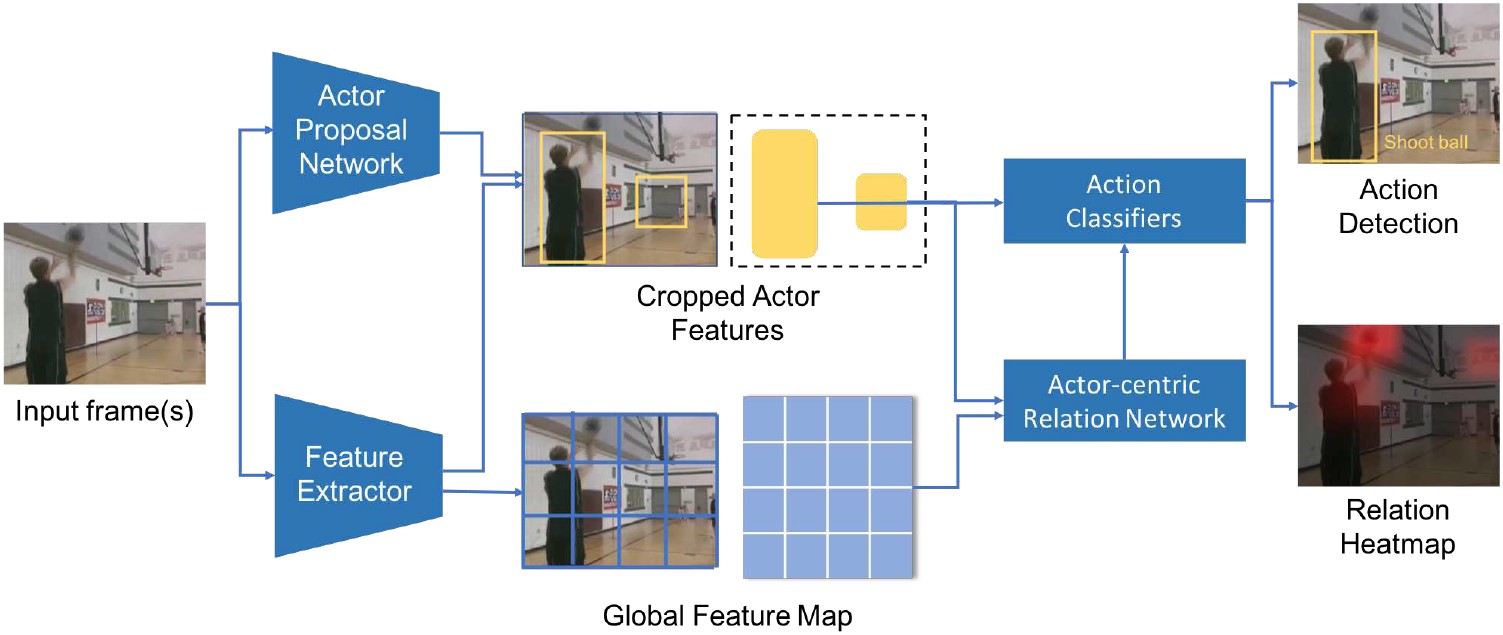}
  \caption{The framework of ACRN~\cite{Sun2018ActorCentricRN}.}
  \label{fig_2018_ACRN}
\end{figure}

\textbf{Visual Relation Modeling}. Detecting and classifying the action of an actor usually depend on its relationships with other actors and objects. Thus, many works strived to model the relationship among actors, objects, and scenes conveyed in the video. For example, \cite{Sun2018ActorCentricRN, ulutan2020actor, jiang2018human, ning2021person, wang2018non, wu2020context, Girdhar2019VideoAT} observed that the surrounding context provides essential information for understanding actions. Accordingly, they proposed a variety of actor-scene interaction models. Concretely, Sun \textit{et al.}~\cite{Sun2018ActorCentricRN} proposed an actor-centric relation network (ACRN) to extract pairwise relations from cropped actor features and a global scene feature, see Fig.~\ref{fig_2018_ACRN}. These relation features are then used for action classification. Wu \textit{et al.}~\cite{wu2020context} empirically found that action recognition accuracy is highly correlated with the resolution of actors. Hence, they cropped and resized image patches around actors and fed them into a 3D CNN to extract actor features that were used to interact with scene features and long-term features.

The attention-based learning strategies are widely adopted in visual relation modeling. For example, non-local network~\cite{wang2018non} leverages self-attention mechanisms to capture long-range dependencies between different entities. PCCA~\cite{ning2021person} utilizes a cross-attention mechanism to model relations between person and context for action detection. Ultan \textit{et al.}~\cite{ulutan2020actor} proposed an actor-conditioned attention maps (ACAM) method that explicitly models the surrounding context and generates features from the complete scene by conditioning them on detected actors. Similarly, Jiang \textit{et al.}~\cite{jiang2018human} designed an actor-target relation network to capture the human-context relationships, which was achieved with a non-local operation between the ROI and its surrounding regions. Girdhar \textit{et al.}~\cite{Girdhar2019VideoAT} proposed an action transformer network that can learn spatio-temporal context from other human actions and objects in a video clip to localize and classify target actions. Their resulting feature embeddings and attention maps were experimentally demonstrated to have a semantic meaning.

There are some works considering multiple interactions. For example, Tang \textit{et al.}~\cite{Tang2020AsynchronousIA} introduced an asynchronous interaction aggregation (AIA) network to explore three kinds of interactions, \textit{i.e.}, person-person, person-object, and temporal interaction. They made them work cooperatively in a hierarchical structure to capture meaningful features.  Zheng \textit{et al.} proposed a network called multi-relation support network (MRSN)~\cite{zheng2023mrsn}, which contains an actor-context relation encoder and an actor-actor relation encoder to model the actor-context and actor-actor relation separately. Then an relation support encoder was deployed to compute the supports between the two relations and performs relation-level interactions. Faure \textit{et al.}~\cite{ faure2022holistic} proposed a bi-modal holistic interaction transformer (HIT) network that comprises an RGB stream and a pose stream. Each stream separately models person, object, and hand interactions. The resulting features from each stream are then glued using an attentive fusion mechanism, and the glued feature is used for action classification. Pramono \textit{et al.} proposed a model called HISAN \cite{pramono2019hierarchical} that combines the two-stream CNNs with hierarchical bidirectional self-attention mechanism to learn the structure relationship among key actors and spatial context to improve the localization accuracy.

Different from the above methods that only consider direct relations between pairs, Pan \textit{et al.}~\cite{Pan2021ActorContextActorRN} proposed an actor-context-actor relation (ACAR) network that takes into account indirect higher-order relations established upon multiple elements. They designed a high-order relation reasoning operator and an actor-context feature bank to fulfil indirect relation reasoning.

Graph neural networks (GNNs) can naturally pass information among entity nodes and model their relations. Thus, GNN is often adopted for visual relation modeling. For instance, Tomei \textit{et al.} proposed a model called STAGE~\cite{tomei2021video} that explores the spatio-temporal relationships through self-attention on a multi-layer graph structure that can connect entities from consecutive clips. Ni \textit{et al.}~\cite{Ni2021IdentityawareGM} proposed an identity-aware graph memory network (IGMN) that highlights the identity information of the actors in terms of both long-term and short-term context, so that the consistency and distinctness between actors are considered for action detection.

Although these frame-level methods have achieved significant results, they do not fully explore temporal continuity as they treat video frames as a set of independent images. Thus, their results can be sub-optimal. To address this issue, clip-level STAD approaches have been proposed, which take as input a sequence of frames and directly output detected tubelet proposals (\textit{i.e.}, a short sequence of bounding boxes).



\subsection{Clip-level Methods}   \label{subsec_clip}

Taking as input a video clip (\textit{i.e.}, a short video snippet), clip-level models directly output 3D spatio-temporal tubelet proposals in this clip. The 3D tubelet proposals are formed by a sequence of bounding boxes that tightly bound the actions of interest. Then, these tubelet proposals in the successive clips are linked together to form complete action tubes. Since being proposed by Jain \textit{et al.}~\cite{jain2014action}, clip-level approaches have become popular among STAD research communities.

Hou \textit{et al.}~\cite{Hou2017TubeCN} presented a tube convolutional neural network (T-CNN) which explores generalizing faster R-CNN~\cite{ren2015faster} from 2D image regions to 3D video volumes. Concerning the RPN and ROI pooling layer of R-CNN, they proposed tube proposal network (TPN) and tube-of-interest (ToI) pooling layer in T-CNN for tubelet generation and spatio-temporal feature pooling, respectively. Kalogeiton \textit{et al.}~\cite{Kalogeiton2017ActionTD} introduced an action tubelet detector (ACT) based on SSD framework~\cite{liu2016ssd}, which takes as input a sequence of frames and then extracts features from each frame with VGG backbone~\cite{simonyan2014very}. These features are stacked to predict scores and regress coordinates for the anchor cuboids to output final action tubelets. Song \textit{et al.}~\cite{Song2019TACNetTC} proposed a transition-aware context network (TACNet) to distinguish transitional states between action and non-action frames so that they could localize the temporal boundaries for the target actions.

\textbf{Large motion}. The aforementioned clip-level methods are all based on an underlying hypothesis that the 3D anchor proposals are cuboid, \textit{i.e.}, having fixed spatial extent across time. Unfortunately, these 3D anchor cuboids can stray far from the flexible ground truth action tubes due to large actor displacement, dramatic actor body shape deformation, large camera motion, \textit{etc.}, particularly for the anchors spanning over a long time. To surpass this limitation, Saha \textit{et al.}~\cite{Saha2017AMTnetAR} proposed a framework that generates two-frame micro-tubes and then links these micro-tubes up into proper action tubes. A year later, they took a further step and explored the anchor micro-tube proposal search space via an approximate transition matrix estimated based on a hidden Markov model (HMM) formulation~\cite{Singh2018TraMNetT}. Their micro-tube hypothesis generation framework can handle large spatial movements in dynamic actors. Four years later, they put forward a new solution to the large-motion case: they proposed to track the actors over time and perform temporal feature aggregation along the respective tracks to enhance actor feature representation~\cite{singh2022spatio}.

He \textit{et al.}~\cite{he2018generic} proposed a method that avoids the 3D cuboid anchor hypothesis by performing frame-level actor detection and then linking the detected bounding boxes to form class-independent action tubelets, which are fed into the temporal understanding module for action classification. Li \textit{et al.}~\cite{he2018generic} proposed a similar framework to \cite{he2018generic}, but they performed STAD in a sparse-to-dense manner: they first generated box proposals at sparsely sampled frames, then they obtained the dense tube by interpolating the sparse proposals across a given detected time interval.

\begin{figure}[t]
  \centering
  \includegraphics[width=3.2in]{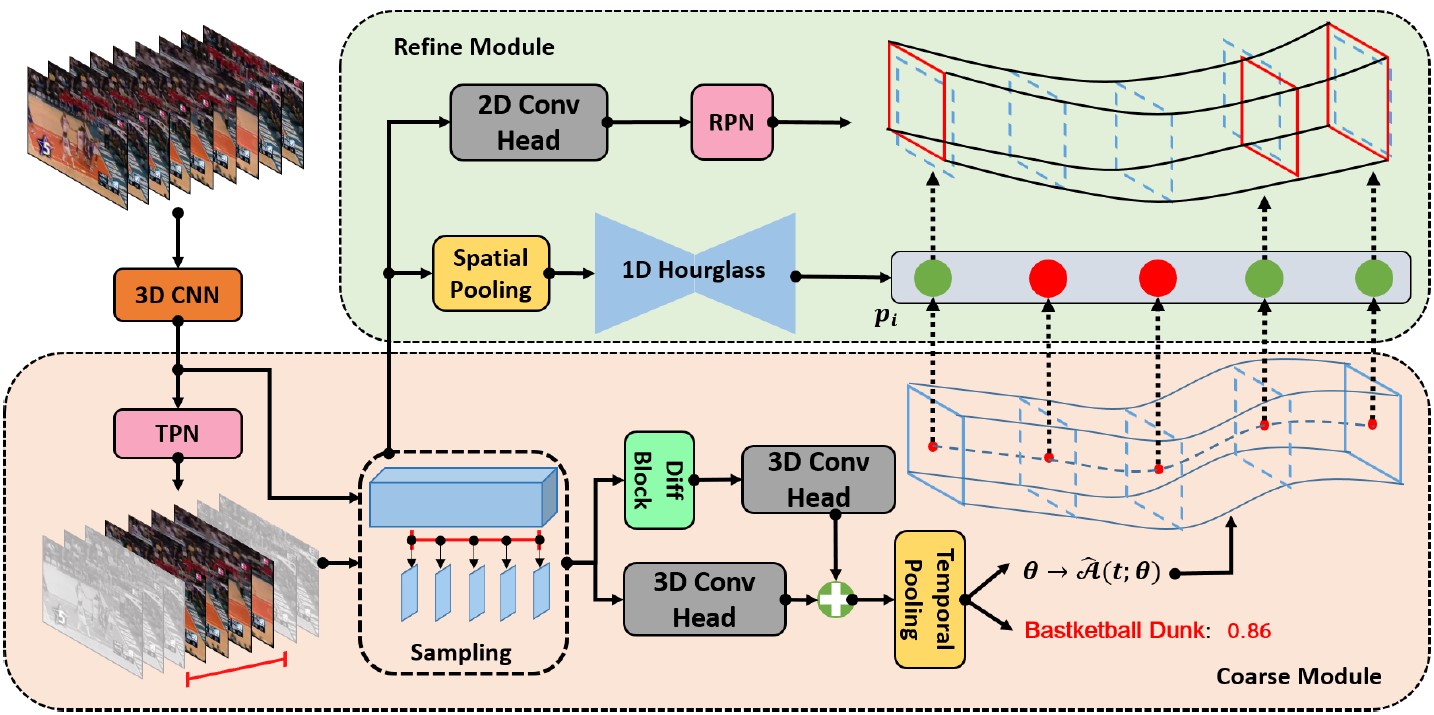}
  \caption{Overview of CFAD framework~\cite{Li2020CFADCA}.}
  \label{fig_2020_CFAD}
\end{figure}

\textbf{Progressive learning}. Unlike the models that perform action detection in one run, progressive learning approaches iteratively refine the proposals towards actions over a few steps. For example, Su \textit{et al.}~\cite{su2019improving} proposed a progressive cross-stream cooperation (PCSC) framework to use region proposals and features from one stream (\textit{i.e.}, Flow/RGB) to help another stream (\textit{i.e.}, RGB/Flow) to improve action localization results in an iterative fashion. Yang \textit{et al.}~\cite{Yang2019STEPSP} proposed a model, named STEP, which progressively refines the pre-defined proposal cuboids across time, \textit{i.e.}, the proposals obtained in current step are fed into the next step for further refinement. Besides, at each step, the 3D proposals are extended in the time dimension to incorporate more adjacent temporal context. Similarly, Li \textit{et al.}~\cite{Li2020CFADCA} proposed a coarse-to-fine action detector (CFAD), which first estimates coarse spatio-temporal action tubes from video streams, and then refines the tubes' location based on key timestamps. As shown in Fig.~\ref{fig_2020_CFAD}, CFAD contains Coarse Module and Refine Module. While the Coarse Module produces initial action tube estimation, the Refine Module selectively adjusts the tube location under the guidance of key timestamps.

\textbf{Anchor free}. The current state-of-the-art methods usually depend on heuristic anchor design and operate on a vast number of pre-defined anchor boxes or cuboids. Although these anchor-based methods have achieved remarkable success, they suffer from critical issues such as huge proposal search space and low efficiency. For example, considering Faster R-CNN~\cite{ren2015faster} for 2D object detection, it requires $KH'W'$ anchors for a feature map with spatial size $H' \times W'$, where usually $K=9$. While for a tubelet across $T'$ frames, the number of 3D anchor proposals soars to $(KH'W')^{T'}$ to maintain the same sampling in space-time. This is a huge number even for a small value of $T'$. To avoid this problem, anchor-free methods for STAD are introduced. For instance, Li \textit{et al.}~\cite{Li2020ActionsAM} viewed each action instance as a trajectory of moving points and presented a model termed as MovingCenter Detector (MOC-detector). It comprises three branches: the center branch for instance center detection and action recognition; the movement branch for movement estimation; the box branch for spatial extent detection. Based on MOC framework, Liu \textit{et al.}~\cite{liu2023accumulated} proposed a lightweight online action detector which encodes short-term action dynamics as accumulated micro-motion. Duarte \textit{et al.}~\cite{duarte2018videocapsulenet} proposed a capsule network for videos, called VideoCapsuleNet, which uses 3D convolutions along with capsules to learn semantic information necessary for action detection and recognition. It has a localization component that utilizes the action representation captured by the capsules for a pixel-wise localization of actions. Recently, inspired by DETR~\cite{carion2020end}, Zhao \textit{et al.}~\cite{zhao2022tuber} proposed a transformer-based framework (termed as TubeR) that directly detects an action tubelet in a video by simultaneously performing action localization and recognition from a single representation. In TubeR, a tubelet-attention module is devised to model the dynamic spatio-temporal nature of a video clip. TubeR learns a set of tubelet queries and outputs action tubelets.

\textbf{Visual Relation Modeling}. Recently, the clip-level (or tubelet-level) visual relations have been explored to enhance STAD models. Li \textit{et al.} introduced a long short-term relation network, dubbed as LSTR~\cite{Li2019LongSR}, which captures both short-term and long-term relations in a video. To be specific, LSTR first produces 3D bounding boxes, \textit{i.e.}, tubelets, in each video clip, and then models short-term human-context interactions within each clip through spatio-temporal attention mechanism and reasons long-term temporal dynamics across video clips via graph convolutional networks in a cascaded manner. In a concurrent work to~\cite{Li2019LongSR}, Zhang \textit{et al.}~\cite{zhang2019structured} proposed an approach where the actors across the video are associated to generate actor tubelets for learning long temporal dependency. The features from actor tubelets and object proposals are then used to construct a relation graph to model human-object manipulation and human-human interaction actions.

\section{Linking up the Detection Results}  \label{sec_link}

Actions are being performed over a period of time, which usually span many frames and clips. After the frame-level or clip-level detection results are obtained, most methods adopt a linking algorithm to link the detections across frames or clips to produce video-level action tubes. In this section, we briefly review the evolution of linking algorithms.

\subsection{Linking up Frame-level Detection Boxes}  \label{sec_link_frame_level}

The first frame-level action detection linking algorithm is introduced by Gkioxari \textit{et al.}~\cite{Gkioxari2014FindingAT}. They assumed that if the spatial extent of two region proposals (\textit{i.e.}, bounding boxes) in adjacent frames have significant overlap and their scores are high, they are highly likely to be linked. Formally, suppose we have two region proposals $R_t$ and $R_{t+1}$ that are located at frame $t$ and frame $t+1$, respectively. For an action class $c$, Gkioxari \textit{et al.}~\cite{Gkioxari2014FindingAT} defined the linking score between these two region proposals to be
\begin{equation}
    \label{equ_score1}
    s_c(R_t, R_{t+1})=s_c(R_t) + s_c(R_{t+1})+\lambda \cdot ov(R_t, R_{t+1}),
\end{equation}
where $s_c(R_i)$ is the class specific score of $R_i$. $ov(R_i, R_j)$ is the IoU of two region proposals $R_i$ and $R_j$. $\lambda$ is a scalar parameter weighting the relative importance of the IoU term. Notably, some models output bounding boxes with associated actionness scores~\cite{Li2018RecurrentTP}, and the class-specific scores in Eq.~(\ref{equ_score1}) are replaced by actionness scores in these works.

After all the linking scores are computed, the optimal path is searched by 
\begin{equation}
    \label{equ_path}
    \bar{R}_c^*=\underset{\bar{R}}{\operatorname{argmax}} \frac{1}{T} \sum_{t=1}^{T-1} s_c(R_t, R_{t+1}),
\end{equation}
where $\bar{R}_c=[R_1, R_2, \ldots, R_T]$ is the sequence of linked regions for action class $c$. This optimization problem is solved via the Viterbi algorithm. After the optimal path is found, the region proposals in $\bar{R}_c^*$ are removed from the set of region proposals and Eq.~(\ref{equ_path}) is solved again~\cite{Gkioxari2014FindingAT}. This process is repeated until the set of region proposals is empty. Each path from Eq.~(\ref{equ_path}) is called an action tube. The score of an action tube $\bar{R_c}$ is defined as $S_c(\bar{R}_c)=\frac{1}{T} \sum_{t=1}^{T-1} s_c(R_t, R_{t+1})$.

Based on the thought of Gkioxari \textit{et al.}~\cite{Gkioxari2014FindingAT}, Peng \textit{et al.}~\cite{Peng2016MultiregionTR} added a threshold function to Eq.~(\ref{equ_score1}) so that the linking score between two region proposals becomes:
\begin{equation}
    \label{equ_score2}
    s_c(R_t, R_{t+1})=s_c(R_t) + s_c(R_{t+1})+\lambda \cdot ov(R_t, R_{t+1}) \cdot \psi(ov),
\end{equation}
where $\psi(ov)$ is a threshold function defined by $\psi(ov)=1$ if $ov$ is larger than $\tau$, $\psi(ov)=0$ otherwise. Peng \textit{et al.}~\cite{Peng2016MultiregionTR} experimentally observed that, with this threshold function, the linking score is better than the one in \cite{Gkioxari2014FindingAT} and more robust due to the additional overlap constraint.

Kopuklu \textit{et al.}~\cite{Kpkl2019YouOW} further extended the linking score definition as follows:
\begin{equation}
    \begin{aligned}
        s_c\left(R_t, R_{t+1}\right)= & \psi(ov) \cdot\left[s_c\left(R_t\right)+s_c\left(R_{t+1}\right)\right. \\
        & +\alpha \cdot s_c\left(R_t\right) \cdot s_c\left(R_{t+1}\right) \\
        & \left.+\beta \cdot o v\left(R_t, R_{t+1}\right)\right],
        \end{aligned}
\end{equation}
where $\alpha$, $\beta$ are scalar parameters. The new term $\alpha \cdot s_c\left(R_t\right) \cdot s_c\left(R_{t+1}\right)$ takes the dramatic change of scores between two successive frames into account and is supposed to improve the performance of video detection in experiments~\cite{Kpkl2019YouOW}.

\textbf{Temporal trimming}. The above linking algorithms produce action tubes spanning the whole video duration. However, human actions usually occupy only a fraction of it. In order to determine the temporal extent of an action instance, some works~\cite{Saha2016DeepLF, Peng2016MultiregionTR} proposed temporal trimming methods. Saha \textit{et al.}~\cite{Saha2016DeepLF} restrained the consecutive proposals to have smooth actionness scores. They solved an energy maximization problem via dynamic programming. Peng \textit{et al.}~\cite{Peng2016MultiregionTR} relied on an efficient maximum subarray method: given a video-level action tube $\bar{R}$, its ideal temporal extent is from frame $s$ to frame $e$ that satisfies the following objective:
\begin{equation}
    s_c(\bar{R}_{(s, e)}^{\star})=\underset{(s, e)}{\operatorname{argmax}}\{\frac{1}{L_{(s, e)}} \sum_{i=s}^e s_c(R_i)-\lambda \frac{|L_{(s, e)}-L_c|}{L_c}\},
\end{equation}
where $L_{(s,e)}$ is the action tube length and $L_c$ is the average duration of class $c$ on the training set.

\textbf{Online action tube generation}. Singh \textit{et al.}~\cite{Singh2016OnlineRM} introduced an online action tube generation algorithm that incrementally (frame by frame) builds multiple action tubes for each action class in parallel. Specifically, for each frame $t$ in a video, the per-class non-maximum suppression (NMS) is conducted to obtain the top $n$ class-specific detection boxes. At the first frame of the video, $n$ action tubes for each class $c$ are initialized by using $n$ detected bounding boxes in this frame. Then, the algorithm grows these action tubes over time by adding one box at a frame or terminates if no matching boxes are found for $k$ consecutive frames~\cite{Singh2016OnlineRM}. Finally, each newly updated tube is temporally trimmed by performing binary labeling using an online Viterbi algorithm.

\subsection{Linking up Clip-level Detection Tubelets}

The clip-level tubelet linking algorithms aim to associate a sequence of clip-level tubelets into video-level action tubes. They are often derived from frame-level box linking algorithms mentioned in the last subsection. For instance, the tubelet linking algorithms in~\cite{Hou2017TubeCN} stems from~\cite{Gkioxari2014FindingAT}, and their intuition is that the content within a tubelet should capture an action
and the connected tubelets in any two consecutive clips should have a large temporal overlap. Therefore, they defined a tubelet's linking score as follows:
\begin{equation}
S=\frac{1}{m} \sum_{i=1}^m { Actionness }_i+\frac{1}{m-1} \sum_{j=1}^{m-1} { Overlap }_{j, j+1}
\end{equation}
where ${ Actionness }_i$ denotes the actionness score of the tubelet from the $i$-th clip. ${ Overlap }_{j, j+1}$ measures the overlap between the linked two proposals respectively from the $j$-th and ($j + 1$)-th clips, and $m$ is the total number of video clips. The overlap between two tubelets is calculated based on the IoU of the last frame of the $j$-th tubelet and the first frame of the $(j+1)$-th tubelet. After computing the tubelets' score, \cite{Hou2017TubeCN} chose a number of linked action tubelets with the highest scores in a video.

In another work, \cite{Kalogeiton2017ActionTD} extended the linking algorithm of~\cite{Singh2016OnlineRM} from frame linking to tubelet linking to build action tubes. Its core idea can be summarized as follows: 
\begin{enumerate}
  \item Initialization: in the first frame of a video, \cite{Kalogeiton2017ActionTD} started a new link for each tubelet. Here, a link refers to a sequence of linked tubelets.
  \item Linking: given a new frame $f$, they extended the existing links with one of the tubelet candidates starting at this frame. They selected the tubelet candidate that met the following criteria: (a) is not already picked by other links, (b) has the highest action score, and (c) its overlap with the link to be extended is higher than a given threshold.
  \item Termination: for an existing link, if the criteria are not met for more than $K$ consecutive frames, the link terminates. $K$ is a given hyperparameter.
\end{enumerate}

Due to its simplicity and efficiency, the tubelet linking algorithm of \cite{Kalogeiton2017ActionTD} was adopted by later works, such as \cite{Li2020ActionsAM, zhao2022tuber}.

\textbf{Temporal trimming}. In the tubelet linking algorithm of \cite{Kalogeiton2017ActionTD}, the initialization and termination steps
determine action tubes' temporal extents, but Song \textit{et al.}~\cite{Song2019TACNetTC} found that it can not thoroughly address the temporal location error induced by transitional state, which is defined as ambiguous states around but not belong to the target actions. To address this issue, Song \textit{et al.}~\cite{Song2019TACNetTC} proposed a transition-aware classifier that can distinguish between transitional states and real actions and thus alleviate the temporal error of spatio-temporal action detection. In follow-up work, Zhao \textit{et al.}~\cite{zhao2022tuber} tried to avoid misclassification for the transitional states by introducing an action switch regression head, which decides whether a box prediction depicts the actor performing the action(s). This regression head gives each bounding box of a tubelet an action switch score. If the score is higher than a given threshold, the box contains an action. Experiments in ~\cite{zhao2022tuber} showed that the action switch regression head could remarkably reduce the misclassification for the transitional states.

\section{Datasets and Evaluation}   \label{sec_dataset}

\subsection{Benchmark Datasets}

Benchmark datasets play significant role in understanding the comparative and absolute strengths and weaknesses of each method. In this section, we first review the commonly used video datasets for STAD, and then we conclude this section by presenting the summary and comparison of these datasets.

\subsubsection{Weizmann}
Weizmann dataset~\cite{blank2005actions} was recorded using a static camera on a uniform background. The actors move horizontally across the frame, maintaining consistency in the actor's size as they perform each action. This dataset comprises a total of 90 video clips grouped 10 action classes, such as ``walking'', ``jogging'', and ``waving'', performed by 9 different subjects. Each video clip contains multiple instances of a single action. The spatial resolution of these videos is 180$\times$144 pixels, and each clip ranges from 1 second to 5 seconds.

\subsubsection{CMU Crowded Videos}
The CMU Crowded Videos~\cite{ke2007event} contains five actions: pick-up, one-hand wave, two-hand wave, push button, and jumping jack. There are 5 training videos for each action and 48 test videos. All videos had been scaled such that the spatial resolution of each video is 120 $\times$ 160. The test videos range from 5 to 37 seconds (166 to 1115 frames). This dataset was recorded with cluttered and dynamic backgrounds so that action detection on this dataset is more challenging than on Weizmann dataset~\cite{blank2005actions}. The CMU Crowded Videos dataset was densely annotated, providing the spatial and temporal coordinates (x, y, height, width, start, and end frames) for specified actions as ground truth.

\subsubsection{MSR Action I and II}
The MSR Action dataset I~\cite{yuan2009discriminative} and II~\cite{cao2010cross} were created by Microsoft Research Group (MSR), and were made publicly available in 2009 and 2010, respectively. The MSR Action dataset II is an extension of I. Whereas the MSR Action dataset I contains 62 action instances in 16 video sequences, the MSR Action dataset II contains 203 instances in 54 videos. 
Each video contains multiple actions performed by different individuals. All videos range from 32 to 76 seconds. Each action instance's spatial and temporal coordinates are provided, allowing the dataset to be used for action detection and recognition.
Both datasets consist of three action classes: clap, hand wave, and boxing. 
Similar to the CMU Crowded dataset~\cite{ke2007event}, the MSR Action datasets were created with cluttered and dynamic backgrounds.

\subsubsection{J-HMDB}
J-HMDB~\cite{jhuang2013towards} is a joint-annotated HMDB~\cite{kuehne2011hmdb} dataset to better understand and analyze the limitations and identify components of algorithms for improvement on overall accuracy on the HMDB dataset. To clarify the description, we first briefly introduce the HMDB dataset and then review the J-HMDB dataset.

HMDB dataset~\cite{kuehne2011hmdb} contain 5 action categories, which are general facial actions (e.g.smile, chew), facial actions with object manipulation (e.g.smoke, eat), general body movements (e.g.cartwheel, clap hands), body movements with object interaction (e.g.brush hair, catch), body movements for human interaction (e.g.fencing, hug). Each category contains a minimum of 101 video clips, and the dataset contains a total of 6849 video clips distributed in 51 action categories.

\begin{table*}[t!]
  \center
  \caption{Summary of major STAD datasets.}
  \begin{tabular}{cccccccc}
    \bottomrule
    Dataset                                    & No. Actions& No. Actors  & No. Videos  & No. Instances& No. Bbox & Year   & Resource   \\
    \hline
    Weizmann~\cite{blank2005actions}           & 10         & 9           & 90          & -            & -        & 2005   & Actor Staged \\
    CMU Crowded Videos~\cite{ke2007event}      & 5          & 6           & 98          & -            & -        & 2007   & Actor Staged \\
    MSR Action I~\cite{yuan2009discriminative} & 3          & 10          & 16          & 62           & -        & 2009   & Actor Staged \\
    MSR Action II~\cite{cao2010cross}          & 3          & >10         & 54          & 203          & -        & 2010   & Actor Staged \\
    UCF Sports~\cite{rodriguez2008action}      & 10         & -           & 150         & -            & -        & 2009   & TV   \\
    J-HMDB~\cite{jhuang2013towards}            & 21         & -           & 928         & 928          & 32k      & 2013   & Movies, YouTube \\
    UCF101-24~\cite{soomro2012ucf101}          & 24         & -           & 3207        & 4458         & 574k     & 2015   & YouTube  \\
    MultiTHUMOS~\cite{yeung2015every}          & 65         & -           & 400         & -            & -        & 2017   & YouTube  \\
    AVA~\cite{Gu2018AVAAV}                     & 80         & -           & 430         & 386k         & 426k     & 2018   & Movies, YouTube  \\
    MultiSports~\cite{li2021multisports}       & 66         & -           & 3200        & 37701        & 902k     & 2021   & YouTube  \\
    \toprule
  \end{tabular}
  \label{table_datasets}
  \end{table*}

J-HMDB~\cite{jhuang2013towards} consists of the selected 21 classes of videos from the HMDB dataset. These selected videos involve a single individual performing the action like brushing hair, jumping, running, \textit{etc}. There are 36 to 55 clips per action class, with each clip containing about 15-40 frames, summing to a total of 928 clips in the dataset. Each clip is trimmed such that the first and last frames correspond to the beginning and end of an action. The frame resolution is 320$\times$240, and the frame rate is 30 fps. 

\subsubsection{UCF Sports}
The UCF Sports \cite{rodriguez2008action} contains 10 actions in the sports domain, which are diving, golf swing, kicking, lifting, horseback riding, running, skateboarding, swinging on a pommel horse and floor, swinging on parallel bars and walking. All the videos contain camera motion and complex backgrounds gathered from the broadcast television channels, such as BBC and ESPN video corpus.
The UCF Sports dataset contains 150 clips and each clip has a frame rate of 10 fps. The spatial resolution of the videos ranges from 480$\times$360 to 720$\times$576 and are 2.20 to 14.40 seconds in duration, averaging 6.39 seconds.

\subsubsection{UCF101-24}
UCF101 dataset \cite{soomro2012ucf101} has been widely used in action recognition research. It comprises realistic videos collected from Youtube containing 101 action categories, with 13320 videos in total. 
UCF101 gives significant diversity in actions with large variations in camera motion, object appearance, viewpoint, cluttered background and illumination conditions, \textit{etc}. 
For the action detection task, a subset of 24 action classes and 3207 videos are provided with dense annotations. This subset is therefore called UCF101-24.
Different from UCF Sports \cite{rodriguez2008action} and J-HMDB \cite{jhuang2013towards}, in which videos are truncated to actions, videos in UCF101-24 are untrimmed.

\subsubsection{THUMOS and MultiTHUMOS }
The THMOS series dataset consists of four datasets: THUMOS'13, THUMOS'14, THUMOS' 15, and MultiTHUMOS. The first three are from THUMOS Challenge took place annually from 2013 to 2015 in conjunction with various major conferences in computer vision \cite{idrees2017thumos, jiang2014thumos, gorban2015thumos, caba2015activitynet}. All of the videos are from UCF101 \cite{soomro2012ucf101}. 
THUMOS datasets consist of 24 action classes. The length of actions varies significantly, \textit{i.e.}, from less than a second to minutes. 
These datasets contain 13,000 temporally trimmed videos, over 1000 temporally untrimmed videos, and over 2500 negative sample videos. These videos might contain none, one, or multiple instances of a single or multiple action(s). MultiTHUMOS~\cite{yeung2015every} is an enhanced version of THUMOS. It is a dense, multi-class, frame-wise labeled video dataset with 400 videos of 30 hours and 38,690 annotations of 65 classes. Averagely, it has 1.5 labels per frame and 10.5 action classes per video.

\subsubsection{AVA}
The Atomic Visual Actions (AVA) dataset is sourced from 430 movies on YouTube. Each movie contributes a clip ranging from the 15th to the 30th minute to the dataset. Each clip is then partitioned into 897 overlapping 3s segments with a stride of 1 second. For each segment, the middle frame is selected as the keyframe. In each keyframe, every person is annotated with a bounding box and (possibly multiple) actions. The 430 movies are split into 235 training, 64 validation, and 131 test movies, roughly a 55:15:30 split, resulting in 211k training, 57k validation, and 118k test segments. AVA dataset contains 80 atomic actions covering pose actions, person-person interactions, and person-object interactions. 60 actions that have at least 25 instances are often adopted for evaluation. This dataset has two annotation versions, \textit{i.e.}, v2.1 and v2.2. The annotation v2.2 is more consistent than v2.1.


\subsubsection{MultiSports}
MultiSports dataset is a new STAD dataset recently released by Li \textit{et al.}~\cite{li2021multisports}. The raw video content of this dataset comes from Olympics and World Cup competitions on YouTube. It consists of 4 sports (\textit{i.e.}, basketball, volleyball, football, and aerobic gymnastics) and 66 action categories. There are 800 clips for each sport, 3200 clips in total. It consists of 37701 action instances with 902k bounding boxes. The instance number of each action category ranges from 3 to 3,477, showing the natural long-tailed distribution. Each video is annotated with multiple instances of multiple action classes. The average video length is 750 frames. Due to the fine granularity of the action labels, the length of each action segment is short, with an average length of 24 frames~\cite{singh2022spatio}.


\subsubsection{Summary and Comparison of STAD Datasets}
\label{compare_datasets}

We summarize the STAD datasets in Table~\ref{table_datasets}, which shows that as research progresses, the datasets supporting STAD studies become increasingly complex in terms of the number of action classes, human subjects, annotation boxes, \textit{etc}. For example, the UCF Sports proposed in 2009 has only 10 action classes and 150 videos, while the MultiSports proposed in 2021 contains 66 actions and 3200 videos. Moreover, in order to keep pace with the growing capability of STAD models, recent datasets are intentionally made increasingly challenging for action detection by adding camera motion, dynamic background, and frequent occlusions. The most recent STAD datasets, such as UCF101-24, AVA, and MultiSports, directly collect videos from YouTube and hence contain videos we would encounter in the real world. Thus, models that perform well in these datasets have great potential for use in real-life scenarios.









\begin{table*}[t!]
  \center
  \caption{Performance comparison on J-HMDB and UCF101-24 datasets. `f.mAP' and `v.mAP' denotes frame-mAP and video-mAP respectively. `-' means that the result is not available. We report the detection accuracy in percentage. The model abbreviations used here refer to the following. I3D: Inflated 3D convolutions~\cite{carreira2017quo}. S3D(+G): Separable 3D convolutions (with gating)~\cite{xie2017rethinking}. NL: Non-local networks~\cite{wang2018non}. 3D-ResNeXt:~\cite{xie2017aggregated}. SF: SlowFast~\cite{feichtenhofer2019slowfast}. DLA34:~\cite{Yu2017DeepLA}.}
  \begin{tabular}{p{0.65cm}<{\centering} p{2.8cm} p{1.7cm}<{\centering} | p{0.8cm}<{\centering} p{0.6cm}<{\centering} p{0.6cm}<{\centering} p{0.6cm}<{\centering} p{0.8cm}<{\centering} | p{0.6cm}<{\centering} p{0.6cm}<{\centering} p{0.6cm}<{\centering} p{0.8cm}<{\centering} p{0.7cm}<{\centering}}
  \bottomrule
  \multirow{3}{*}{Year}   & \multirow{3}{*}{Method} & \multirow{3}{*}{Backbone}   & \multicolumn{5}{c|}{J-HMDB}  & \multicolumn{5}{c}{UCF101-24} \\
  \cline{4-8} \cline{9-13}
    &        &            & \multirow{2}{*}{f.-mAP} & \multicolumn{4}{c|}{v.-mAP} & \multirow{2}{*}{f.-mAP}      & \multicolumn{4}{c}{v.-mAP} \\
  \cline{5-8} \cline{10-13}
    &        &                                                                 &      & @0.2 & @0.5 & @0.75& 0.5:0.95 &   & @0.2  & @0.5  & @0.75& 0.5:0.95\\
  \hline
  2015   & Gkioxari \textit{et al.}\cite{Gkioxari2014FindingAT}     & AlexNet  & 36.2 &  -   & 53.3 &  -   &  -   &  -    &  -    &  -    &  -   & -\\
  2015   & Weinzaepfel \textit{et al.}\cite{Weinzaepfel2015LearningTT}& AlexNet& 45.8 & 63.1 & 60.7 &  -   &  -   & 35.8  & 51.7  &  -    &  -   & -\\
  2016   & MR-TS R-CNN\cite{Peng2016MultiregionTR}                  & VGG      & 58.5 & 74.3 & 73.1 &  -   &  -   & 65.7  & 72.9  &  -    &  -   & -\\
  2016   & H-FCN \cite{Wang2016ActionnessEU}                        & H-FCN    & 39.9 &  -   & 56.4 &  -   &  -   &  -    &  -    &  -    &  -   & -\\
  2016   & Saha \textit{et al.} \cite{Saha2016DeepLF}               & VGG      &  -   & 72.6 & 71.5 & 43.3 & 40.0 &  -    & 66.7  & 35.9  & 7.9  & 14.4\\
  2017   & ACT \cite{Kalogeiton2017ActionTD}                        & VGG      & 65.7 & 74.2 & 73.7 & 52.1 & 44.8 & 67.1  & 77.2  & 51.4  & 22.7 & 25.0\\
  2017   & Singh \textit{et al.}\cite{Singh2016OnlineRM}            & VGG      &  -   & 73.8 & 72.0 & 44.5 & 41.6 &  -    & 73.5  & 46.3  & 15.0 & 20.4\\
  2017   & Zolfaghari \textit{et al.}~\cite{zolfaghari2017chained}  & C3D      &  -   & 78.2 & 73.5 &  -   &  -   &  -    & 47.6  & 26.8  &  -   & -\\
  2017   & AMTnet \cite{Saha2017AMTnetAR}                           & VGG      &  -   &  -   &  -   &  -   &  -   &  -    & 79.4  & 51.2  & 19.0 & 23.4\\
  2017   & CPLA \cite{Yang2017SpatioTemporalAD}                     & VGG      &  -   &  -   &  -   &  -   &  -   &  -    & 73.5  & 37.8  &  -   & -\\
  2017   & T-CNN \cite{Hou2017TubeCN}                               & C3D      & 61.3 & 78.4 & 76.9 &  -   &  -   & 41.4  & 47.1  &  -    &  -   & -\\
  2018   & Gu \textit{et al.}\cite{Gu2018AVAAV}                     & I3D      & 73.3 &  -   & 78.6 &  -   &  -   & 76.3  &  -    & 59.9  &  -   & -\\
  2018   & ACRN \cite{Sun2018ActorCentricRN}                        & S3D-G    & 77.9 &  -   & 80.1 &  -   &  -   &  -    &  -    &  -    &  -   & -\\
  2018   & RTPR \cite{Li2018RecurrentTP}                            & Res101   &  -   & 82.7 & 81.3 &  -   &  -   &  -    & 77.9  &  -    &  -   & -\\
  2018   & TraMNet \cite{Singh2018TraMNetT}                         & VGG      &  -   &  -   &  -   &  -   &  -   &  -    & 79.0  & 50.9  & 20.1 & 23.9\\
  2018   & TPN \cite{he2018generic}                                 & VGG      &  -   & 79.7 & 77.0 &  -   &  -   &  -    & 71.7  &  -    &  -   & -\\
  2018   & Alwando \textit{et al.}\cite{alwando2018cnn}             & VGG      &  -   & 79.9 & 78.3 &  -   &  -   &  -    & 72.9  & 41.1  &  -   & -\\
  2018   & VideoCapsuleNet \cite{duarte2018videocapsulenet}         & -        & 64.6 & 95.1 &  -   &  -   &  -   & 78.6  & 97.1  & 80.3  &  -   & -\\
  2019   & LSTR \cite{Li2019LongSR}                                 & Res101   &  -   & 86.9 & 85.5 &  -   &  -   &  -    & 83.0  & 64.4  &  -   & -\\
  2019   & HISAN \cite{pramono2019hierarchical}                     & Res101   &  -   & 87.6 & 86.5 & 53.8 & 51.3 &  -    & 82.3  & 51.5  & 23.5 & 24.9\\
  2019   & STEP \cite{Yang2019STEPSP}                               & VGG      &  -   &  -   &  -   &  -   &  -   & 75.0  & 76.6  &  -    &  -   & -\\
  2019   & TACNet \cite{Song2019TACNetTC}                           & VGG      & 65.5 & 74.1 & 73.4 & 52.5 & 44.8 & 72.1  & 77.5  & 52.9  & 21.8 & 24.1\\
  2019   & 2in1 \cite{zhao2019dance}                                & VGG      &  -   &  -   & 74.7 & 53.3 & 45.0 &  -    & 78.5  & 50.3  & 22.2 & 24.5\\
  2019   & PCSC \cite{su2019improving}                              & I3D      & 74.8 & 82.6 & 82.2 & 63.1 & 52.8 & 79.2  & 84.3  & 61.0  & 23.0 & 27.8\\
  2019   & YOWO \cite{Kpkl2019YouOW}                                &3D-ResNeXt& 74.4 & 87.8 & 85.7 & 58.1 &  -   & 87.2  & 75.8  & 48.8  &  -   & -\\
  2020   & MOC \cite{Li2020ActionsAM}                               & DLA34    & 70.8 & 77.3 & 77.2 & 71.7 & 59.1 & 78.0  & 82.8  & 53.8  & 29.6 & 28.3\\
  2020   & AIA \cite{Tang2020AsynchronousIA}                        & SF101    &  -   &  -   &  -   &  -   &  -   & 78.8  &  -    &  -    &  -   & -\\
  2020   & C-RCNN \cite{wu2020context}                              & Res50-I3D& 79.2 &  -   &  -   &  -   &  -   &  -    &  -    &  -    &  -   & -\\
  2020   & CFAD \cite{Li2020CFADCA}                                 & I3D      &  -   & 86.8 & 85.3 &  -   &  -   & 72.5  & 81.6  & 64.6  &  -   & 26.7\\
  2020   & ACAM \cite{ulutan2020actor}                              & I3D      & 78.9 &  -   & 83.9 &  -   &  -   &  -    &  -    &  -    &  -   & -\\
  2020   & Li \textit{et al.}\cite{li2020finding}                   & I3D      &  -   & 76.1 & 74.3 & 56.4 &  -   &  -    & 71.1  & 54.0  & 21.8 & -\\
  2021   & ACDnet \cite{Liu2021ACDnetAA}                            & VGG      &  -   &  -   &  -   &  -   &  -   & 70.9  &  -    &  -    &  -   & -\\
  2021   & ACAR-Net \cite{Pan2021ActorContextActorRN}               & SF101    &  -   &  -   &  -   &  -   &  -   & 84.3  &  -    &  -    &  -   & -\\
  2021   & SAMOC \cite{Ma2021SpatioTemporalAD}                      & DLA34    & 73.1 & 79.2 & 78.3 & 70.5 & 58.7 & 79.3  & 80.5  & 53.5  & 30.3 & 28.7\\
  2021   & WOO \cite{Chen2021WatchOO}                               & SF101    & 80.5 &  -   &  -   &  -   &  -   &  -    &  -    &  -    &  -   & -\\
  2022   & SE-STAD \cite{Sui2022ASA}                                & SF101-NL & 82.5 &  -   &  -   &  -   &  -   &  -    &  -    &  -    &  -   & -\\
  2022   & TAAD \cite{singh2022spatio}                              & SF50     &  -   &  -   &  -   &  -   &  -   & 81.5  &  -    &  -    &  -   & -\\
  2023   & MRSN \cite{zheng2023mrsn}                                & SF50     & -    &  -   &  -   &  -   &  -   & 80.3  &  -    &  -    &  -   & -\\  
  2022   & TubeR \cite{zhao2022tuber}                               & I3D      &  -   & 81.8 & 80.7 &  -   &  -   & 81.3  & 85.3  & 60.2  &  -   & 29.7\\
  2023   & HIT \cite{faure2022holistic}                             & SF50     & 83.8 & 89.7 & 88.1 &  -   &  -   & 84.8  & 88.8  & 74.3  &  -   & -\\
  2023   & EVAD \cite{chen2023efficient}                            & ViT-B    & 90.2 & -    & -    &  -   &  -   & 85.1  & -     & -     &  -   & -\\
  \toprule
  \end{tabular}
  \label{table_JHMDB_UCF}
  \end{table*}

\begin{table*}[t!]
\center
\caption{Performance comparison on UCF Sports and MultiSports datasets.}
\begin{tabular}{clc|ccccc|ccc}
  \bottomrule
  \multirow{3}{*}{Year}  & \multirow{3}{*}{Method}  & \multirow{3}{*}{Backbone}  & \multicolumn{5}{c|}{UCF Sports}  & \multicolumn{3}{c}{MultiSports}\\
  \cline{4-11}
                        &                          &                            & \multirow{2}{*}{f.-mAP}  & \multicolumn{4}{c|}{v.-mAP}  & \multirow{2}{*}{f.-mAP}  & \multicolumn{2}{c}{v.-mAP}\\
  \cline{5-8} \cline{10-11}
                          &                         &                          &       & @0.2  & @0.5  & @0.75 & 0.5:0.95 &       & @0.2  & @0.5    \\
  \hline
  2015   & Gkioxari \textit{et al.}\cite{Gkioxari2014FindingAT}              & AlexNet  & 68.1  &  -    & 75.8  &  -    &  -       &  -    &  -    & \\
  2015   & Weinzaepfel \textit{et al.}\cite{Weinzaepfel2015LearningTT}       & AlexNet  & 71.9  &  -    & 90.5  &  -    &  -       &  -    &  -    & \\
  2016   & MR-TS R-CNN\cite{Peng2016MultiregionTR}                  & VGG      & 84.5  & 94.8  & 94.7  &  -    &  -       &  -    &  -    & \\
  2016   & H-FCN \cite{Wang2016ActionnessEU}                        & H-FCN    & 82.7  &  -    &  -    &  -    &  -       &  -    &  -    & \\
  2017   & ACT \cite{Kalogeiton2017ActionTD}                        & VGG      & 87.7  & 92.7  & 92.7  & 78.4  & 58.5     &  -    &  -    & \\
  2017   & T-CNN \cite{Hou2017TubeCN}                               & C3D      & 86.7  & 95.2  &  -    &  -    &  -       &  -    &  -    & \\
  2018   & RTPR \cite{Li2018RecurrentTP}                            & Res101   &  -    & 98.6  & 98.6  &  -    &  -       &  -    &  -    & \\
  2018   & TPN \cite{he2018generic}                                 & VGG      &  -    & 96.0  & 95.7  &  -    &  -       &  -    &  -    & \\
  2018   & Alwando \textit{et al.}\cite{alwando2018cnn}                      & VGG      &  -    & 94.7  & 94.7  & 89.6  & 67.5     &  -    &  -    & \\
  2018   & VideoCapsuleNet \cite{duarte2018videocapsulenet}         & -        & 83.9  & 97.1  &  -    &  -    &  -       &  -    &  -    & \\
  2019   & LSTR \cite{Li2019LongSR}                                 & Res101   &  -    & 98.9  & 98.9  &  -    &  -       &  -    &  -    & \\
  2019   & SlowFast \cite{feichtenhofer2019slowfast}                & SF101    &  -    &  -    &  -    &  -    &  -       & 27.2  &  24.2 & 9.7\\
  2019   & 2in1 \cite{zhao2019dance}                                & VGG      &       &       & 92.7  & 83.4  &  -       &       &       & \\
  2019   & YOWO \cite{Kpkl2019YouOW}                                & 3D-ResNeXt-101 &  -    &  -    &  -    &  -    &  -       & 9.3   & 10.8  & 0.9\\
  2020   & MOC \cite{Li2020ActionsAM}                               & DLA34    &  -    &  -    &  -    &  -    &  -       & 25.2  & 12.9  & 0.6\\
  2020   & CFAD \cite{Li2020CFADCA}                                 & I3D      &  -    & 94.5  & 92.7  &  -    &  -       &  -    &  -    & \\
  2020   & Li \textit{et al.}\cite{li2020finding}                            & I3D      &  -    & 94.3  & 93.8  & 79.5  &  -       &  -    &  -    & \\
  2022   & TAAD \cite{singh2022spatio}                              & SF50     &  -    &  -    &  -    &  -    &  -       & 55.3  & 60.6  & 37.0\\
  2023   & HIT \cite{faure2022holistic}                             & SF50     &  -    &  -    &  -    &  -    &  -       & 33.3  & 27.8  & 8.8\\
  \toprule
\end{tabular}
\label{table_UCFsports_MultiSports}
\end{table*}

\subsection{Evaluation Metrics}

The performances of STAD methods are often evaluated by two popular metrics, \textit{i.e.}, frame and video mean Average Precision (mAP), which are usually denoted as frame-mAP and video-mAP. 

\subsubsection{Frame-mAP}
Frame-mAP measures the area under the precision-recall curve of the bounding box detections at each frame. A detection is correct if its IoU with the ground truth bounding box is larger than a given threshold and the action label is correctly predicted~\cite{Gkioxari2014FindingAT}. The threshold is often set as 0.5. Frame-mAP allows researchers to compare the detection accuracy independently of the linking strategy.

\subsubsection{Video-mAP}
Video-mAP measures the area under precision-recall curve of the action tube predictions. A tube detection is correct if its IoU with the ground truth tube is larger than a given threshold and the action label is correctly predicted~\cite{Gkioxari2014FindingAT}. The IoU between two tubes is defined as the IoU over the temporal domain, multiplied by the average of the IoU between boxes averaged over all overlapping frames~\cite{Weinzaepfel2015LearningTT}. The threshold for video-mAP is often set as 0.2, 0.5, 0.75, and 0.5:0.95, corresponding to the average video-mAP for thresholds with step 0.05 in this range. Whereas frame-mAP measures the ability of classification and spatial detection in a single frame, video-mAP can further evaluate the performance of temporal detection.

\subsection{Performance Analysis}

In this section, we analyze the performance of the state-of-the-art methods for STAD. Since J-HMDB, UCF101-24, UCF Sports, MultiSports, and AVA datasets are the most commonly used in deep learning-based STAD, we report the performance of state-of-the-art methods on these five datasets.

Table~\ref{table_JHMDB_UCF} presents the performance of STAD methods on the J-HMDB and UCF101-24 datasets. VideoCapsuleNet~\cite{duarte2018videocapsulenet} achieved the best results in terms of video-mAP. VideoCapsuleNet is different from the mainstream STAD methods. Whereas the mainstream STAD methods usually involve box or tubelet proposal generation and linking these proposals, VideoCapsuleNet is a 3D capsule network that performs pixel-wise action segmentation along with action classification~\cite{duarte2018videocapsulenet}. This is probably why VideoCapsuleNet is excluded by most state-of-the-art methods for performance comparison. Apart from VideoCapsuleNet, the transformer-based frameworks, \textit{e.g.}, HIT~\cite{faure2022holistic}, TubeR~\cite{zhao2022tuber} and EVAD~\cite{chen2023efficient}, achieved the best performance, demonstrating the capability of transformers in STAD. This is further verified by the results in Table~\ref{table_UCFsports_MultiSports} and Table~\ref{table_AVA}. 

Table~\ref{table_UCFsports_MultiSports} provides the performance of STAD methods on the UCF Sports and MultiSports datasets. The MultiSports dataset was constructed very recently, so the reported results on this dataset are fewer than other datasets. Besides, the performance on this dataset is relatively low. This is probably because MultiSports action instances have large actor displacement, and there are often multiple actions in one video, making the detection difficult.

Table~\ref{table_AVA} shows the results of STAD methods on the AVA (v2.1 and v2.2) dataset. The remarkable performance of AIA~\cite{Tang2020AsynchronousIA}, ACAR-Net~\cite{Pan2021ActorContextActorRN}, IGMN~\cite{Ni2021IdentityawareGM} and HIT~\cite{faure2022holistic} indicates that visual relation and long-term temporal context is critical for the model to learn discriminative representation and can improve the detection accuracy.

\section{Future Directions}   \label{sec_future}

In this section, we discuss the future directions in STAD that might be interesting to explore.

\begin{table*}[h!]
  \center
  \caption{Performance comparison on AVA dataset under the metric of frame-mAP with threshold of 0.5. The column `Flow' denotes whether optical flow is used as input.}
  \begin{tabular}{clcccccc}
    \bottomrule
    Year   & Method                                                   & Backbone   & Flow        & Detector                     & Pretrained  & AVA (v2.1)  & AVA (v2.2)\\
    \hline
    2018   & Gu \textit{et al.}\cite{Gu2018AVAAV}                              & I3D        &\Checkmark   & Faster R-CNN~\cite{ren2015faster} & Kinetics-400&15.8         &- \\
    2018   & ACRN \cite{Sun2018ActorCentricRN}                        & S3D-G      &\Checkmark   & Faster R-CNN~\cite{ren2015faster} & Kinetics-400&17.4         &- \\
    2018   & RTPR \cite{Li2018RecurrentTP}                            & Res101     &\Checkmark   & -                                 & -           &22.3         & \\
    2018   & Girdhar \textit{et al.} \cite{girdhar2018better}                  & I3D        &\XSolidBrush & Faster R-CNN~\cite{ren2015faster} & Kinetics-400&21.9         &- \\
    2019   & LSTR \cite{Li2019LongSR}                                 & Res101     &\XSolidBrush & -                                 & -           &25.3         &- \\
    2019   & SlowFast \cite{feichtenhofer2019slowfast}                & SF101      &\XSolidBrush & Faster R-CNN~\cite{ren2015faster} & Kinetics-600&28.2         & \\
    2019   & STEP \cite{Yang2019STEPSP}                               & I3D        &\XSolidBrush & -                                 & Kinetics-400&18.6         &- \\
    2019   & Girdhar \textit{et al.}\cite{Girdhar2019VideoAT}                  & I3D        &\XSolidBrush & -                                 & Kinetics-400&24.9         & \\
    2019   & LFB \cite{wu2019long}                               & Res101-I3D-NL   &\XSolidBrush & Faster R-CNN~\cite{ren2015faster} & Kinetics-400&27.7         & \\
    2019   & Zhang \textit{et al.}\cite{zhang2019structured}                   & I3D        &\XSolidBrush & Dave \textit{et al.} \cite{dave2019towards}& Kinetics-400&22.2         &- \\
    2020   & AIA \cite{Tang2020AsynchronousIA}                        & SF101      &\XSolidBrush & Faster R-CNN~\cite{ren2015faster} & Kinetics-700&31.2         &34.4 \\
    2020   & C-RCNN \cite{wu2020context}                          & Res50-I3D-NL   &\XSolidBrush & Faster R-CNN~\cite{ren2015faster} & Kinetics-400&28.0         & \\
    2020   & X3D \cite{feichtenhofer2020x3d}                          & X3D-XL     &\XSolidBrush & Faster R-CNN~\cite{ren2015faster} & Kinetics-600&27.4         & \\
    2020   & ACAM \cite{ulutan2020actor}                              & I3D        &\XSolidBrush & Faster R-CNN~\cite{ren2015faster} & Kinetics-400&24.4         & \\
    2021   & ACAR-Net \cite{Pan2021ActorContextActorRN}               & SF101      &\XSolidBrush & Faster R-CNN~\cite{ren2015faster} & Kinetics-400&30.0         &- \\
    2021   & IGMN \cite{Ni2021IdentityawareGM}                        & SF101      &\XSolidBrush & Faster R-CNN~\cite{ren2015faster} & Kinetics-700&30.2         &33.0 \\
    2021   & WOO \cite{Chen2021WatchOO}                               & SF101      &\XSolidBrush & Sparse R-CNN~\cite{sun2021sparse} & Kinetics-600&28.0         &28.3 \\
    2021   & STAGE \cite{tomei2021video}                              & SF101      &\XSolidBrush & Faster R-CNN~\cite{ren2015faster} & Kinetics-600&29.8         & \\
    2022   & SE-STAD \cite{Sui2022ASA}                                & SF101-NL   &\XSolidBrush & FCOS~\cite{tian2019fcos}          & Kinetics-600&28.8         &29.3 \\
    2022   & TAAD \cite{singh2022spatio}                              & SF50       &\XSolidBrush & YOLOv5~\cite{Brostom2020real}     & Kinetics-700&31.8         &- \\
    2022   & TubeR \cite{zhao2022tuber}                               & CSN-152    &\XSolidBrush & -                                 &IG+Kinetics-400&32.0       &33.6 \\
    2023   & MRSN \cite{zheng2023mrsn}                                & SF101      &\XSolidBrush & Faster R-CNN~\cite{ren2015faster} & Kinetics-400&-            &33.5 \\
    2023   & HIT \cite{faure2022holistic}                             & SF101      &\XSolidBrush & Faster R-CNN~\cite{ren2015faster} & Kinetics-700&-            &32.6 \\
    2023   & EVAD \cite{chen2023efficient}                            & ViT-B      &\XSolidBrush & -                                 & Kinetics-400&31.1         &32.2 \\
    \toprule
  \end{tabular}
  \label{table_AVA}
  \end{table*}

\textbf{Label-efficient learning for STAD}. The annotation for STAD contains not only the action class but also the bounding boxes of the actors and the start and end of action instances, making the collection and annotation of data expensive and time-consuming. To alleviate this burden, some pioneering works, such as~\cite{zhang2022semi, arnab2022beyond}, have been introduced. While \cite{zhang2022semi} proposed a semi-supervised framework to utilize the unlabeled video actions, \cite{arnab2022beyond} presented a co-finetuning method to leverage the large-scale action classification datasets like Kinetics~\cite{kay2017kinetics} and Something-Something v2~\cite{goyal2017something}. Nevertheless, much more effort is needed to explore the label-efficient STAD algorithm.

\textbf{Online real-time STAD}. STAD has numerous online applications, such as surveillance and autonomous driving. In these applications, the system must process the newly captured frame or clip only based on history and the current data and report the detection result as soon as possible. To reach this goal, the model is required to be lightweight and efficient. This is a challenging task. Although \cite{dave2022gabriellav2, Singh2016OnlineRM} have taken the first step to undertake this task, it is still far from fully explored, and more effort is needed.
  
\textbf{STAD under large motion}. In real scenarios, many actions are with large motion because of fast actor displacement, rapid camera motion, \textit{etc}. STAD under large motion is an interesting direction worth exploring.

\textbf{Multimodal learning for STAD}. On the one hand, action video naturally contains multiple modalities, including visual, acoustic, and even linguistic messages (\textit{e.g.}, caption). Thus, fully utilizing these modalities via multimodal learning for STAD has the potential to achieve better detection accuracy than single-modality learning. On the other hand, actions can be captured by various sensors, such as depth camera, infrared camera, inertial sensor, and LiDAR. STAD might benefit from the fusion of the representations learned from these multimodal data.

\textbf{Diffusion models for STAD}. Diffusion models~\cite{song2020denoising, ho2020denoising, song2019generative, song2020score} are a class of generative models that starts from the sample in random distribution and recover the data sample via a gradual denoising process. They have recently become one of the hottest topics in computer vision. Even though they belong to generative model, they are freshly demonstrated effective for representative perception
tasks, such as object detection~\cite{chen2022diffusiondet} and temporal action location~\cite{nag2023difftad, liu2023diffusion, xu2023boundary}. Taking as input random spatial boxes (or temporal proposals), the diffusion-based models can yield object boxes (or action proposals) accurately. Since STAD can be regarded as the combination of object detection and temporal action location, these pioneering works~\cite{chen2022diffusiondet, nag2023difftad, liu2023diffusion, xu2023boundary} shed light on leveraging diffusion models for solving STAD task.

\section{Conclusion}   \label{sec_conclusion}

Expedited by the rapid advances of deep neural networks, spatio-temporal action detection has made significant progress in recent years. This survey has extensively reviewed deep learning-based spatio-temporal action detection methods from different aspects, including models, datasets, linking algorithms, performance comparison, and future directions. The comparative summary of methods, datasets, and performance in pictorial and tabular forms clearly shows their attributes which will benefit the interested researchers. We hope this comprehensive survey will foster further research in spatio-temporal action detection.

\appendix

\end{CJK}
\bibliographystyle{CVMbib}
\bibliography{manuscript}

\end{document}